\documentclass{article}

\usepackage{graphicx}
\usepackage{amsthm}
\theoremstyle{plain}

\theoremstyle{definition}

\usepackage{wrapfig}
\usepackage{algorithm}
\usepackage{algorithmic}
\usepackage{tabularray}
\usepackage{xcolor}
\usepackage{amssymb}
\usepackage{amsmath}

\PassOptionsToPackage{numbers,compress}{natbib}
\usepackage[preprint]{neurips_2026}

\usepackage[utf8]{inputenc} 
\usepackage[T1]{fontenc}    
\usepackage{hyperref}       
\usepackage{url}            
\usepackage{booktabs}       
\usepackage{amsfonts}       
\usepackage{nicefrac}       
\usepackage{microtype}      
\usepackage{xcolor}         
\usepackage{multirow}

\title{PAT-VCM: Plug-and-Play Auxiliary Tokens for Video Coding for Machines}

\author{
  Wei Jiang \\
  Futurewei Technologies Inc.\\
  San Jose, CA 95131 \\
  \texttt{wjiang@futurewei.com} \\
 \And
   Wei Wang \\
  Futurewei Technologies Inc. \\
 San Jose, CA 95131 \\
 \texttt{rickweiwang@futurewei.com} \\
}

\begin{document}

\maketitle

\begin{abstract}
Existing video coding for machines is often trained for a specific downstream task and model. As a result, the compressed representation becomes tightly coupled to the end task, making it difficult to scale across multiple tasks or adapt to model updates. We propose PAT-VCM, a plug-and-play auxiliary-token framework for video coding for machines. PAT-VCM keeps a shared baseline compressed stream and augments it with lightweight task-aware auxiliary tokens, allowing different downstream tasks to recover the information they need by training lightweight auxiliary modules instead of a separate full codec for each task. The framework supports visual residual tokens and prompt/control tokens in the main pipeline, and can also accommodate semantic tokens as an additional extension. We evaluate PAT-VCM on detection, segmentation and depth estimation. A shared detection-oriented auxiliary branch provides a reusable first refinement, task-specific visual branches improve segmentation and depth, and prompt tokens provide further segmentation gains at negligible bitrate. These results suggest that a shared compressed representation, combined with lightweight task-aware auxiliary tokens, is a practical and scalable alternative to tightly task-coupled VCM design.

\end{abstract}

\section{Introduction}

Learned visual compression has made steady progress through end-to-end optimized autoencoders, hierarchical priors, autoregressive entropy models, and neural video codecs \cite{agustsson2020ssf,balle2017e2e,balle2018hyperprior,lu2020dvc,minnen2018joint}. More recently, tokenized and generative approaches have started to represent video with discrete or latent tokens decoded by strong generative models \cite{agarwal2025cosmos,tvc}. These methods improve compactness and reconstruction quality, but their main objective is still visual reconstruction for human viewing.

In many deployment settings, compressed video is consumed not only by humans but also by machine models for detection, segmentation, depth estimation, etc. This has led to growing interest in coding for machines, task-oriented compression, and joint human--machine coding, where the goal is not only to preserve pixel quality but also to preserve downstream task performance \cite{chamain2021machine,gao2023taskgeneric,wood2022taskoriented,zhang2024allinone}. In this setting, the question is not just how well a bitstream reconstructs pixels, but whether it preserves the information needed by different machine tasks under a limited bitrate.

Most machine-oriented compression methods still assume that one compressed representation can support all downstream tasks. In practice, this is difficult for two reasons. First, different tasks rely on different information. Detection depends on object-level localization cues. Segmentation is more sensitive to boundaries and decoder interaction. Depth estimation depends on local geometric structure. A single  token stream is therefore unlikely to preserve all of these equally well. Second, current VCM methods are often trained for a specific downstream task model, which makes the compressed representation tightly coupled to the end task used during training. Such a design is hard to scale: each new task may require retraining a new codec, and even different downstream models for the same task may favor different task-specific representations. As a result, it is difficult to support heterogeneous machine tasks with one shared compressed stream.

This motivates the main idea of our paper. Instead of forcing one compressed representation to serve all downstream tasks, we start from a shared baseline codec and augment it with \emph{auxiliary tokens}. These tokens are task-aware additions to the baseline bitstream. They are used to recover or preserve information that is important for a downstream task but not well retained in the shared compressed representation. Different auxiliary streams can have different token types, including visual residual tokens, prompt/control tokens, and semantic tokens.

Based on this idea, we propose \textbf{PAT-VCM}, a plug-and-play auxiliary-token framework for video coding for machines. PAT-VCM consists of a shared baseline codec and attachable auxiliary token streams. The baseline codec produces a primary compressed stream and a coarse reconstructed video that are shared across downstream tasks. When a task needs information that is not preserved well enough by this shared stream, an auxiliary branch can be trained and attached on top. This keeps the baseline representation shared while moving task adaptation to lightweight auxiliary modules.

A key feature of PAT-VCM is that the auxiliary design is modular. The auxiliary streams are attached to the baseline codec, rather than requiring a separate full codec for each task. The downstream models remain frozen and operate on refined decoded video, decoder-side prompts, or transmitted semantic description. This keeps the method at the codec level and makes it easier to support new tasks or model updates without retraining the full system.

We evaluate PAT-VCM on several downstream tasks with different requirements. A shared detection-oriented auxiliary stream improves localization from compressed video and provides a reusable first refinement. On top of it, a segmentation-specific visual stream and a compact point-prompt codebook improve SAM-based mask prediction \cite{kirillov2023sam,ravi2024sam2}. A depth-specific visual stream improves monocular depth estimation under a frozen depth backbone \cite{yang2024depthanything}. Together, these results show that a shared compressed representation can be made substantially more useful for heterogeneous machine tasks by adding lightweight task-aware auxiliary tokens.

Our contributions are as follows:
\begin{itemize}
    \item We introduce PAT-VCM, a plug-and-play auxiliary-token framework for video coding for machines, which is built on a shared baseline compressed stream with attachable task-aware auxiliary streams.
    \item We present a unified auxiliary-token view of machine-oriented compression, covering visual residual tokens, prompt/control tokens, and semantic tokens.
    \item We show that a shared detection-oriented auxiliary stream can serve as a reusable early refinement across downstream tasks, while later auxiliary branches remain task-specific.
    \item We validate the framework on representative tasks including segmentation and depth estimation, and show that different auxiliary token types address different downstream bottlenecks under modest bitrate overhead.
\end{itemize}

\section{Related Work}

\subsection{Learned, Tokenized, and Generative Visual Compression}

Learned image and video compression has progressed through end-to-end optimized transform coding, hierarchical priors, autoregressive entropy models, and neural predictive video codecs \cite{agustsson2020ssf,balle2017e2e,balle2018hyperprior,lu2020dvc,minnen2018joint}. More recently, tokenized and generative codecs have explored discrete or latent visual representations decoded by strong neural generators, shifting part of reconstruction from hand-designed predictive structure to learned priors over visual content \cite{agarwal2025cosmos,generativevcmsurvey2024}. These methods provide increasingly compact and expressive compressed representations, but remain centered on reconstruction quality, whether measured by distortion or perceptual realism. Our work starts from the same tokenized-compression setting, but asks a different question: how should a shared compressed representation be augmented when decoded video is primarily consumed by downstream machine models?

\subsection{Scalable and Layered Compression}

 In classical scalable video coding, a base layer provides a coarse representation and enhancement layers progressively improve reconstruction quality, such as the scalable extension of H.264/AVC \cite{schwarz2007svc}. More recent learned compression work has also explored progressive or scalable bitstreams through layered or staged refinement \cite{hojjat2023progdtd,zheng2024deepfgs}. PAT-VCM shares the high-level structure of a shared baseline stream plus additional refinement streams, but differs in purpose. Traditional scalable coding is reconstruction-oriented, whereas PAT-VCM uses auxiliary streams to recover information needed by downstream machine tasks. The auxiliary streams are therefore task-aware rather than fidelity-aware, and can target different bottlenecks such as visual evidence or decoder control.

\subsection{Coding for Machines and Task-Oriented Compression}

Coding for machines and task-oriented compression aim to preserve downstream task performance under rate constraints rather than optimizing only human perceptual quality \cite{chamain2021machine,wood2022taskoriented}. Existing work has explored feature-preserving compression, semantic communication, task-aware image coding, and joint human--machine coding, typically targeting one downstream task or one shared machine representation \cite{gao2023taskgeneric,zhang2024allinone}. Our work is aligned with this motivation, but differs in formulation. Instead of learning one task-specific codec or assuming that one compressed representation should serve all downstream tasks equally well, PAT-VCM introduces a shared baseline codec together with task-aware auxiliary token streams. This shifts the focus from single-task optimization to modular support for heterogeneous downstream tasks.

\subsection{Frozen Downstream Models and Promptable Decoders}

Recent vision models provide strong frozen models for a range of downstream tasks. DETR offers an end-to-end transformer formulation for object detection \cite{carion2020detr}. Promptable segmentation models such as SAM and SAM~2 expose decoder-side conditioning through boxes and points, making them especially relevant for studying control signals under compression \cite{kirillov2023sam,ravi2024sam2}. For geometry, MiDaS and Depth Anything provide robust monocular depth estimators that generalize well across datasets \cite{ranftl2021midas,yang2024depthanything}. Our PAT-VCM is designed for this setting: the downstream models remain frozen, while the codec is augmented with auxiliary tokens that refine either the decoded video itself or the control signal presented to the downstream decoder.

\subsection{Shared Representations and Task-Specific Adaptation}

Multi-task learning often combines a shared representation with lightweight task-specific adaptation \cite{caruana1997multitask}. Similar ideas appear in residual adapters for vision, where a shared backbone is augmented by small task-dependent modules \cite{rebuffi2017residual}, and in prompt-based adaptation, where task-specific control is introduced without changing the backbone itself \cite{jia2022visualprompt}. Our setting is related, but the adaptation happens at the codec level rather than inside the downstream model. In PAT-VCM, the baseline compressed stream serves as the shared representation, while auxiliary tokens provide task-specific correction or control for different tasks.

\section{Method}

\begin{figure}[t]
\centering
\includegraphics[width=\textwidth]{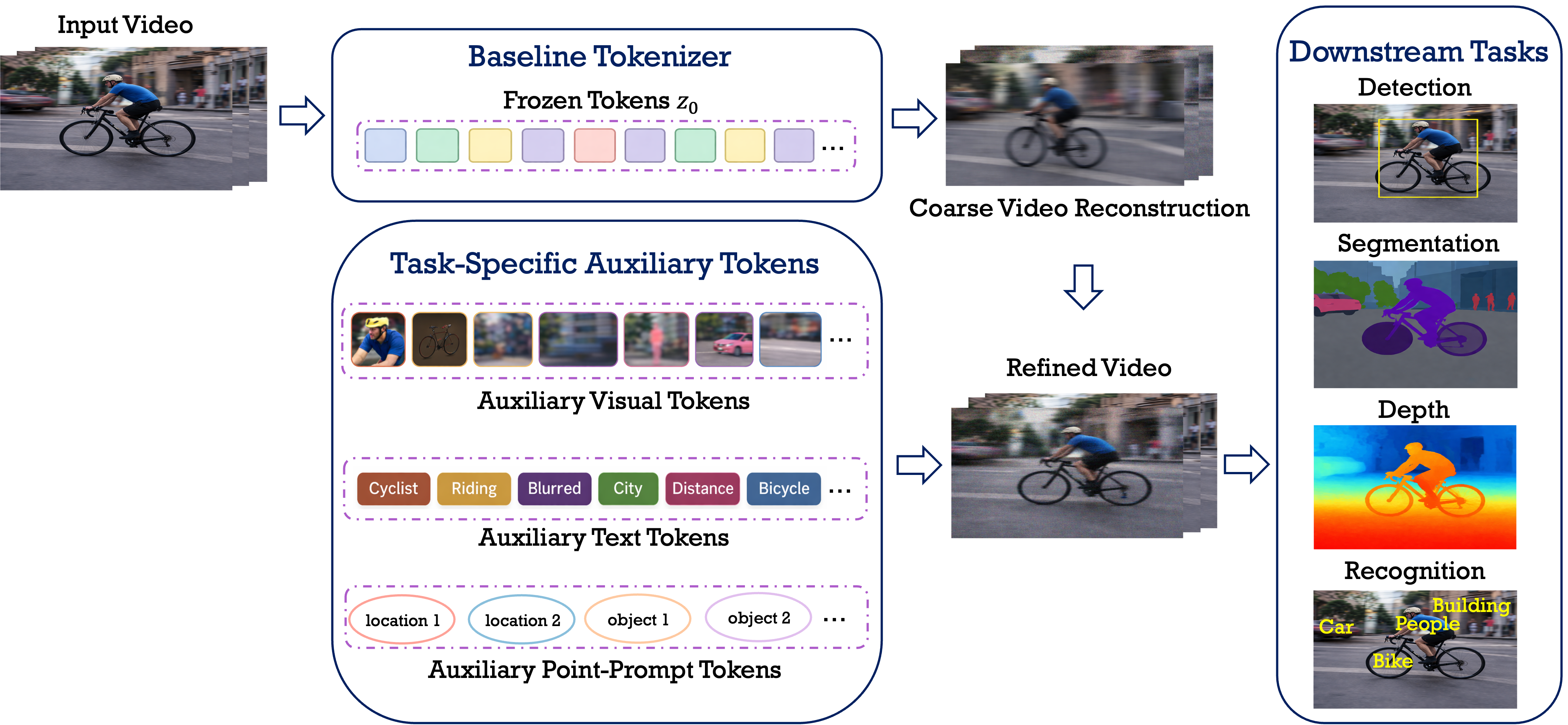}\vspace{-.5em}
\caption{Overview of PAT-VCM. A shared baseline stream provides a common compressed representation and coarse reconstruction. On top of this baseline, auxiliary streams can be attached to preserve or recover information needed by different downstream tasks. These auxiliary streams may take different forms, including visual residual tokens, text-based tokens, and prompt/control tokens.}
\label{fig:overview}
\end{figure}

\subsection{PAT-VCM Formulation}
\label{sec:method}

The basic observation is that one compressed representation is rarely sufficient for many downstream machine tasks. Detection, segmentation and depth estimation do not rely on exactly the same information. A bitstream that is good for general reconstruction may still lose cues that matter for a particular task.

PAT-VCM addresses this by separating the codec into two parts: a shared baseline stream and task-aware auxiliary streams (as illustrated in Figure \ref{fig:overview}). The baseline codec produces a compact primary representation and a coarse reconstructed video that are shared across tasks. Auxiliary streams are then attached only when a downstream task needs additional information that is not preserved well enough by the baseline representation.

Let $x \in \mathbb{R}^{B \times c \times T \times H \times W}$ denote an input video, where \(B\), \(c\), \(T\) are the batch size, number of channels, and number of frames, and \(H\) and \(W\) are the spatial resolution. The baseline codec consists of an encoder \(E\), a quantizer \(Q\), and a decoder \(D\):
\[
z_0 = Q(E(x)), \qquad \hat{x}_0 = D(z_0),
\]
where \(z_0\) is the primary compressed stream and \(\hat{x}_0\) is the corresponding coarse reconstructed video.

The baseline codec is trained with a standard rate--distortion objective,
\[
\mathcal{L}_{\text{base}} = \mathcal{L}_{\text{recon}} + \beta \mathcal{L}_{\text{rate}} + \lambda \mathcal{L}_{\text{quant}},
\]
where \(\mathcal{L}_{\text{recon}}\) is reconstruction loss, \(\mathcal{L}_{\text{rate}}\) is bitrate loss, and \(\mathcal{L}_{\text{quant}}\) is the quantization regularization loss. After training, the baseline codec is frozen and used as a shared compressed representation.

For a downstream task \(t\), PAT-VCM introduces a task-specific auxiliary stream \(a_t\). In the visual-auxiliary case, the auxiliary stream is produced by an auxiliary encoder applied to the selected ROI crop from the baseline compressed representation together with the corresponding ROI crop from the current reconstructed video:
\[
a_t = E_t^{\mathrm{aux}}(z_0|_{R_t}, \hat{x}_0|_{R_t}).
\]
It is then decoded into an ROI residual,
\[
r_t = G_t(a_t,\hat{x}_0), \qquad \hat{x}_t = \hat{x}_0 + r_t,
\]
where \(G_t\) is a task-specific refinement decoder and \(r_t\) denotes a pixel-space residual written back only within the selected ROI regions \(R_t\).

More generally, the auxiliary stream does not have to act through pixel refinement. Let \(c_t\) denote the task conditioning derived from the auxiliary stream, so that the downstream model takes the form
\[
y_t = f_t(\hat{x}_t, c_t).
\]
For visual auxiliary streams, \(c_t=\varnothing\) and the auxiliary information is expressed through the refined video \(\hat{x}_t\). For prompt/control tokens, \(c_t\) denotes decoder-side prompts such as points or boxes. For semantic tokens, \(c_t\) denotes compact task-level semantics transmitted directly by the encoder. This gives a unified view in which the primary stream \(z_0\) remains shared, while the auxiliary stream either refines the decoded video or provides task-specific conditioning for the downstream model.

A deliberate design choice in PAT-VCM is that the downstream models remain frozen. This keeps the adaptation at the codec level. It also avoids retraining a new codec together with every new downstream model. This is important in the long run because downstream models evolve quickly in practice: if each task or model update required codec retraining, deployment and maintenance would become expensive and difficult to scale.

\subsection{Auxiliary Token Types}

PAT-VCM supports different types of auxiliary tokens because different downstream tasks lose different kinds of information under compression.

The first type is a \emph{visual auxiliary stream}. These tokens are decoded into a pixel-space residual and are used when the task suffers from missing local detail, degraded boundaries, or weakened region-level structure.

The second type is a \emph{prompt/control stream}. These tokens do not primarily change the pixels. Instead, they modify how a frozen downstream decoder is queried. The point-prompt codebook in the  segmentation setting is an example of such tokens.

The third type is a \emph{semantic token stream}. These tokens transmit lightweight task-level semantics computed by the encoder on the original frame. They are useful when the downstream task can benefit more from compact semantic information than from further pixel refinement. In PAT-VCM, semantic tokens are supported as an additional extension of the framework and are studied in Appendix~\ref{ap:seg_texttoken}, \ref{ap:poseestimation}, \ref{ap:recognitiontask} for tasks such as segmentation, semantic recognition and pose estimation. 

\subsection{Representative Task Families}

All task-specific visual branches share the same basic design: an ROI is processed by an auxiliary encoder, quantized, and decoded into a ROI-localized pixel-space residual that is written back to the reconstructed video. The architecture is shared across tasks, and only the supervision changes. This keeps the framework modular, since adding a new visual task does not require a whole new codec, only a frozen downstream model and a task-specific loss.

PAT-VCM is meant to be a general auxiliary-token framework rather than a task-specific pipeline. We therefore study several representative task families built on the same shared baseline codec.

\paragraph{Detection.}
Detection provides the shared early refinement. Starting from the coarse reconstructed video \(\hat{x}_0\), the codec refines ROI regions that are likely to affect localization and feeds the refined video to the same frozen detector. After training, this detection-oriented branch is frozen and reused by all later task-specific branches.

\paragraph{Segmentation.}
For segmentation, we use a task-specific visual auxiliary branch together with an optional prompt branch. The visual branch is trained to improve the input seen by a frozen SAM-based segmentor. We further use a compact point-prompt codebook to refine decoder interaction. The codebook contains \(32\) candidate point locations, so transmitting one selected point takes \(5\) bits per ROI, while a foreground-background pair takes \(10\) bits per ROI. This setting combines two complementary mechanisms: visual refinement improves the reconstructed input, while prompt refinement improves the decoder query. We also study a text-based semantic token variant in Appendix~\ref{ap:seg_texttoken}, where encoder-side text can provide additional gains on hard segmentation cases with adaptive transmission.

\paragraph{Depth estimation.}
Depth estimation is a dense geometric prediction task. As in segmentation, the same auxiliary architecture is applied on ROI regions, but with a different supervision signal. The depth-specific auxiliary branch is trained so that the refined video produces depth predictions closer to those from the original frame under a frozen monocular depth model. This shows that the same auxiliary-token framework extends beyond mask prediction to absolute per-pixel regression.

\subsection{Additional Tasks in the Appendix}

We report several additional task studies in the appendix, including semantic recognition \ref{ap:recognitiontask}, pose estimation \ref{ap:poseestimation}, and surface normal estimation \ref{ap:surfacenormal}. These studies help clarify both the broader applicability and the current limits of the framework. In particular, semantic recognition and pose estimation illustrate semantic-token use cases, where compact semantics can be transmitted directly at negligible bitrate. Surface normal estimation, on the other hand, shows that some tasks benefit less under the current ROI-based residual design.

\section{Experiments}

\subsection{Experimental Setup}

We use the frozen Cosmos DV4x8x8 tokenizer and reconstruction model \cite{agarwal2025cosmos} as the baseline codec. The downstream output from this compressed video is denoted as \emph{Cosmos}, and the output from the uncompressed original video is denoted as \emph{Orig}. 

For downstream tasks, we use frozen models throughout: Deformable DETR for detection \cite{carion2020detr}, SAM-based segmentation \cite{kirillov2023sam,ravi2024sam2}, and Depth Anything V2 Small for depth estimation \cite{yang2024depthanything}.

All visual auxiliary branches share the same core architecture: an auxiliary encoder, an FSQ module with levels \([8,8,8,8]\), and a residual decoder that predicts a pixel-space correction for selected ROIs. The shared detection-oriented branch (\textbf{Det-Aux}) is applied first, followed by task-specific branches such as segmentation-oriented \textbf{Seg-Aux} and depth-oriented \textbf{Depth-Aux}. For segmentation, we also evaluate a point-prompt codebook as a prompt/control stream.

ROI boxes are selected by the frozen Deformable DETR. For Det-Aux, we use detections on the Cosmos reconstruction with confidence in \([0.05,0.5)\), keep at most 3 boxes per frame, and expand them by a factor of 1.3. For later task-specific branches, we re-run the same detector on the Det-Aux-enhanced frame, keep detections with confidence at least 0.3, retain at most 3 boxes per frame, and expand them by a factor of 2.0. Encoder and decoder have the same detector, so the ROI set is reproduced without transmitting box coordinates.

For segmentation, we evaluate on DAVIS \cite{perazzi2016davis,ponttuset20172017}, MOSE \cite{ding2023mose}, and VIPSeg \cite{miao2022vipseg}, and report mean IoU. For depth estimation, we evaluate on DAVIS and VIPSeg. Following standard monocular depth evaluation, we use per-frame scale--shift alignment and report AbsRel and \(\delta<1.25\) \cite{eigen2015predicting,eigen2014depth}, for both full-frame and ROI-only regions.

All visual auxiliary branches share the same architecture and differ only in task-specific supervision. Det-Aux uses detector-feature distillation and an ROI-localized reconstruction loss. Seg-Aux uses segmentation-model feature distillation inside ROI regions together with an ROI-localized reconstruction loss. Depth-Aux uses a scale-invariant log-depth loss under a frozen depth model plus an ROI-localized reconstruction regularizer. Full loss definitions, weights, auxiliary branch train/eval split, and training configuration details are provided in Appendix~\ref{ap:trainingobjectives}.

\subsection{Shared Detection-Oriented Auxiliary Stream}

We begin with the shared detection-oriented branch, Det-Aux. Its role is to recover object-level cues weakened by compression and to provide stronger region hypotheses for later tasks. On DAVIS, Det-Aux improves detection recall from \(0.516\) to \(0.549\) and matched detection IoU from \(0.475\) to \(0.512\) over the Cosmos baseline. Full detection details are provided in Appendix~\ref{ap:detection}.

\subsection{Segmentation}

\begin{table*}[t]
\centering
\caption{Full segmentation results. Metric is mean IoU.}
\label{tab:seg_full}\vspace{-.1em}
\begin{tabular}{lcccccc}
\toprule
Dataset & Orig & Cosmos & Det-Aux & + Seg-Aux & + Seg-Aux + 1-pt & + Seg-Aux + FG+BG\\
\midrule
DAVIS  & 0.774 & 0.677 & 0.686 & 0.709 & 0.751 & \textbf{0.764}  \\
MOSE   & 0.713 & 0.693 & 0.694 & 0.707 & 0.743 & \textbf{0.753} \\
VIPSeg & 0.731 & 0.721 & 0.723 & 0.727 & 0.751 & \textbf{0.754}  \\
\bottomrule
\end{tabular}\vspace{-.7em}
\end{table*}

\paragraph{Segmentation-specific visual refinement.}
We first apply a task-specific visual branch, Seg-Aux, on top of Det-Aux. The downstream segmentor remains frozen, so Seg-Aux can improve segmentation only by improving the visual evidence seen by the segmentor. 

\paragraph{Prompt/control tokens.}
We further use a compact prompt/control stream. A prompt token is a discrete index into a fixed codebook of candidate point locations inside the Det-Aux ROI box. The codebook contains \(32\) candidates, constructed from a \(6\times 6\) grid with the four corners removed. At the encoder, where the original frame is available, we first compute a target mask by running SAM-huge (641M parameters) \cite{kirillov2023sam} on the original frame with the ROI box. We then evaluate all candidate points in the codebook by running SAM-base (93.7M parameters) \cite{kirillov2023sam} on the refined compressed frame using the same ROI box and the candidate point as a prompt, and comparing the resulting mask with the target mask from the original frame. The candidate that achieves the highest mask IoU is selected, and only its codebook index is transmitted. At the decoder, this index is mapped back to a point location inside the ROI box and used as a prompt for SAM-base. A one-point prompt therefore requires only \(5\) bits per ROI. We also consider a foreground-background variant, in which a second background point is selected by the same search procedure conditioned on the chosen foreground point, requiring \(10\) bits per ROI in total.

\begin{figure}[t]
\centering
\includegraphics[width=\textwidth]{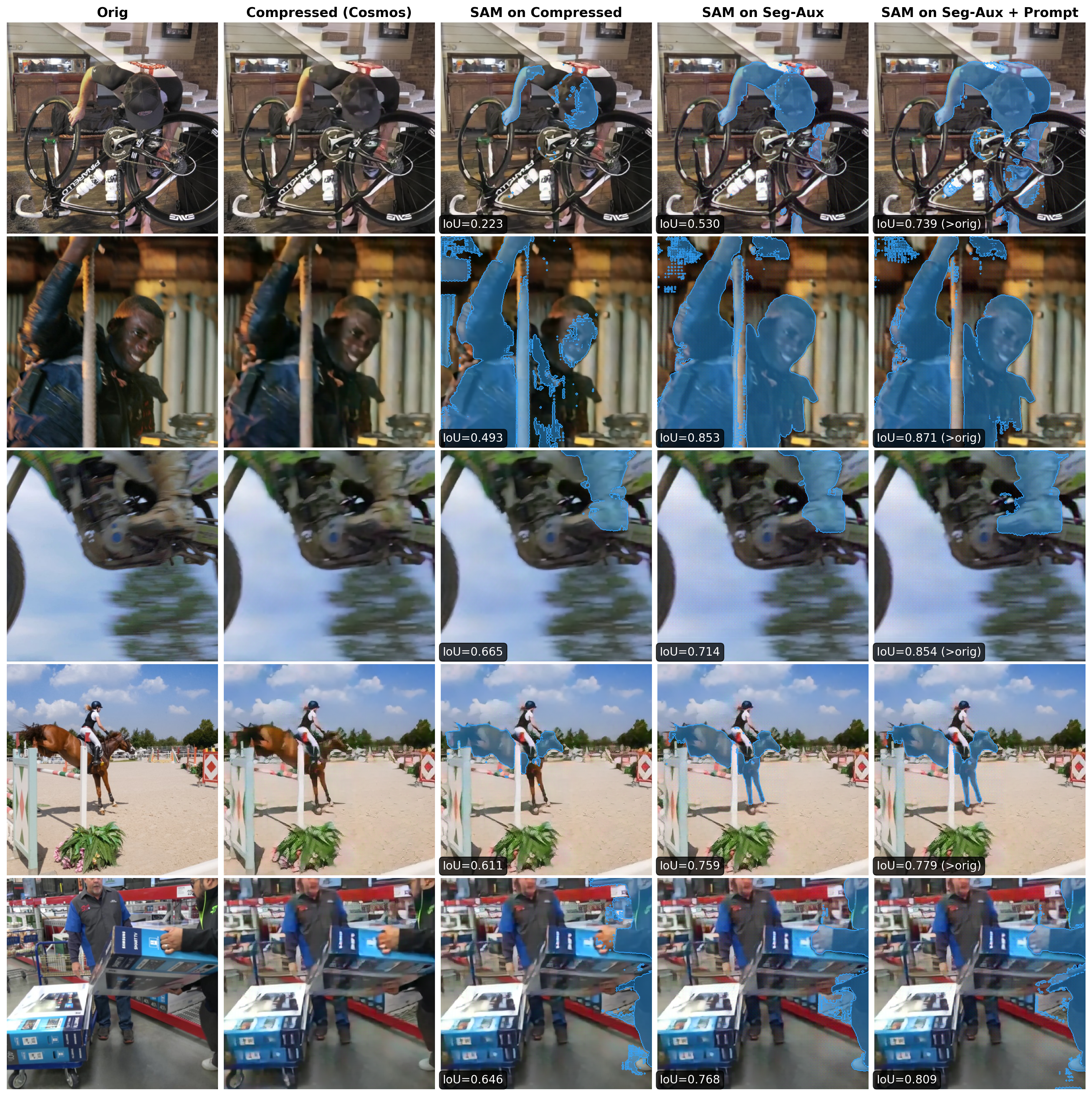}\vspace{-.5em}
\caption{Qualitative segmentation results. From left to right: original frame, compressed reconstruction, segmentation on the compressed reconstruction, result after Seg-Aux, and result after Seg-Aux plus prompt tokens. Examples marked with ``(>Orig)'' exceed the segmentation quality obtained from the uncompressed original frame under the same frozen SAM-based evaluation.}
\label{fig:seg_qual}\vspace{-1em}
\end{figure}

\paragraph{Full segmentation results.}
Table~\ref{tab:seg_full} summarizes the full segmentation results. On DAVIS, the full chain closes most of the compression gap relative to the original video. On MOSE and VIPSeg, the final result exceeds the output obtained from the original video under the same frozen segmentor. Figure~\ref{fig:seg_qual} shows representative qualitative examples. Several examples are marked with ``(>Orig)'', indicating that the refined compressed representation yields better segmentation than the uncompressed original video under the same frozen SAM-based evaluation.

\paragraph{Text-based semantic tokens.}
We further evaluated a text-based semantic token for segmentation. On MOSE val, text tokens generated at the encoder from the uncompressed source are particularly helpful on hard ROIs, improving IoU from \(0.098\) to \(0.246\) when the Cosmos+Seg-Aux performance is below \(0.30\). Their effect is not uniformly positive across all ROIs, but under an adaptive transmission policy that sends text only for hard cases, the overall IoU improves from \(0.682\) to \(0.692\) with negligible bitrate overhead. Full details are given in Appendix~\ref{ap:seg_texttoken}.

\paragraph{Additional segmentation analysis.}
We defer several additional segmentation analyses to Appendix~\ref{ap:segmentation}, including a token-type ablation, a difficulty-bin analysis on DAVIS, an ROI-size sensitivity study, and a no-harm retention check on easy cases. Overall, these analyses show that auxiliary refinement provides the largest gains on medium and hard cases, that the strongest improvements are obtained on medium and large ROIs, and that combining visual and prompt tokens is more effective than using either one alone.

\subsection{Depth Estimation}

\paragraph{Depth-specific visual refinement.}
Depth-Aux uses the same auxiliary architecture as the segmentation branch, but is trained with a depth-specific objective. The main loss is a scale-invariant log-depth loss on ROI regions, together with an L1 regularizer between the refined ROI and the Det-Aux ROI. The pseudo-ground-truth depth is obtained by applying the frozen depth model to the original frame.

\begin{figure*}[t]
\centering
\includegraphics[width=\textwidth]{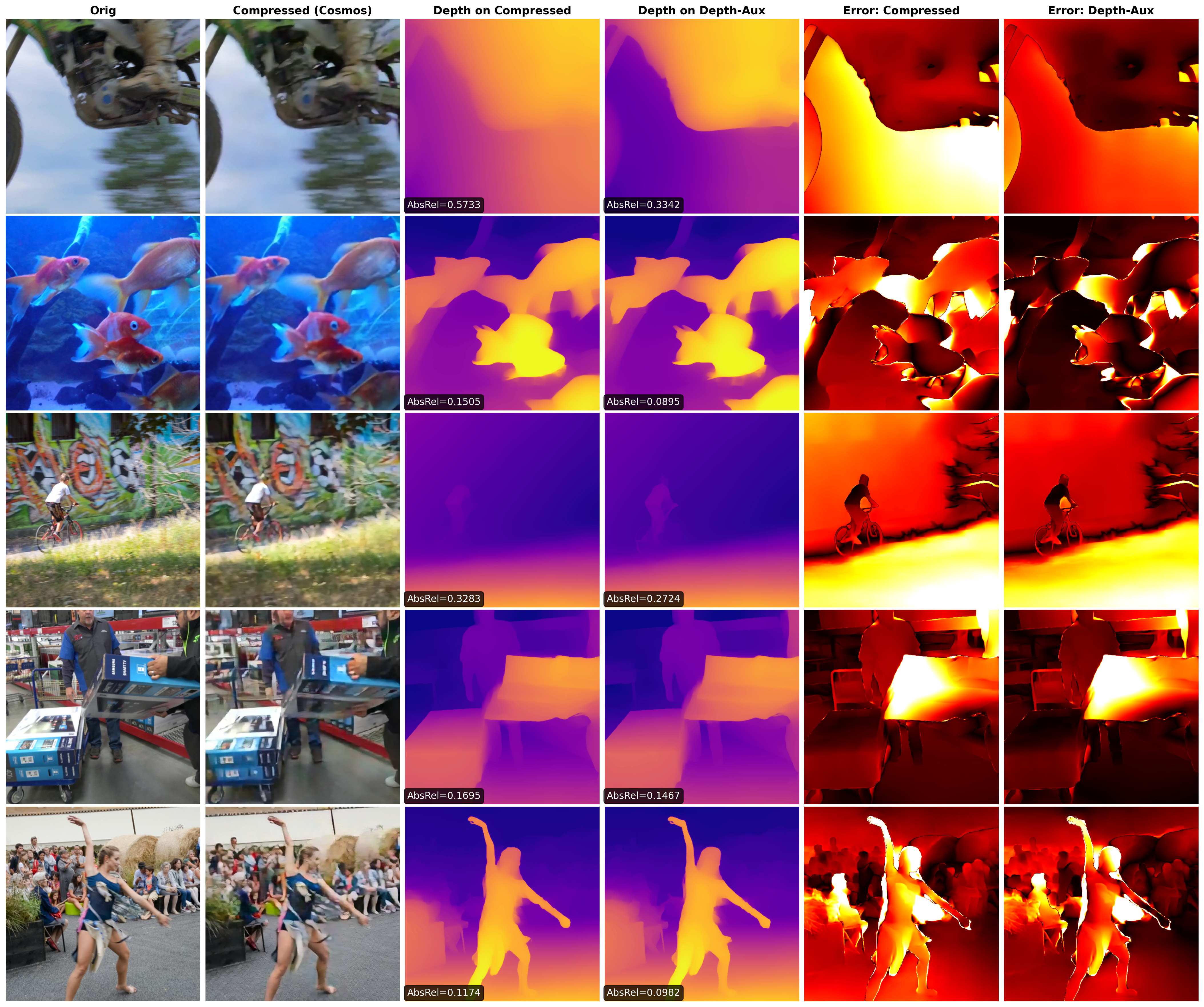}\vspace{-.5em}
\caption{Qualitative depth estimation results. From left to right: original frame, compressed reconstruction, depth from the compressed reconstruction, result after Depth-Aux, error map for the compressed reconstruction, and error map for the refined result. Examples marked with ``(>Orig)'' exceed the depth quality obtained from the uncompressed original frame.}
\label{fig:dpt_qual}\vspace{-.7em}
\end{figure*}

Table~\ref{tab:depth_davis} reports the DAVIS results. Depth-Aux is highly effective, reducing ROI AbsRel from \(3.764\) to \(1.554\), corresponding to a \(59\%\) reduction relative to Cosmos. Figure~\ref{fig:dpt_qual} shows qualitative examples. Compared with the compressed baseline, Depth-Aux recovers more coherent object-level depth structure and reduces large local errors inside ROI regions. Several examples are marked with ``(>Orig)'', indicating that the refined compressed representation yields better depth estimation than the uncompressed original under the same frozen Depth Anything V2 evaluation protocol.

\begin{table}[t]
\centering
\caption{Depth auxiliary results on DAVIS. AbsRel is lower-is-better, \(\delta<1.25\) is higher-is-better.}
\label{tab:depth_davis}\vspace{-.3em}
\begin{tabular}{lccc}
\toprule
Variant & Full AbsRel & ROI AbsRel & ROI \(\delta<1.25\) \\
\midrule
Cosmos & 2.576 & 3.764 & 0.828 \\
+ Det-Aux & 2.151 & 3.057 & 0.834 \\
+ Det-Aux + Depth-Aux & \textbf{1.290} & \textbf{1.554} & \textbf{0.840} \\
\bottomrule
\end{tabular}
\end{table}

\paragraph{Additional depth analysis.}
We defer several additional depth analyses to Appendix~\ref{ap:depthestimation}. These include a difficulty-bin analysis and an ROI-size sensitivity study. These analyses show that the largest gains come from difficult cases and larger ROIs, while easy cases change only modestly.

\subsection{Rate-Matched Evaluation}

To separate the effect of the auxiliary architecture from the effect of additional bits alone, we compare PAT-VCM against a frozen Cosmos CV4 quantization sweep, which provides a rate-matched baseline within the same model family. Cosmos CV4 is a continuous-latent variant \cite{agarwal2025cosmos}, and its scalar quantization level can be varied to obtain different operating bitrates. We also compare against MPA~\cite{liu2024mpa}, a recent joint human--machine image codec. Full details are given in Appendix~\ref{ap:isorate}.

On DAVIS segmentation, the full PAT-VCM pipeline (Cosmos + Det-Aux + Seg-Aux + FG+BG) reaches \(0.764\) IoU. Since the FG+BG prompt adds only \(10\) bits per ROI, its bitrate overhead is negligible relative to the visual auxiliary stream, so this result is effectively at the same operating point as Cosmos + Det-Aux + Seg-Aux. At this rate, PAT-VCM substantially outperforms the nearest higher-rate Cosmos CV4 reference point (\(0.509\) IoU at \(0.167\) bpp) and also exceeds MPA under the matched SAM protocol. These results show that the gain does not come from added rate alone, but from how auxiliary information is allocated between visual refinement and decoder control.

\begin{table*}[t]
\centering
\caption{Rate-matched evaluation on DAVIS. Segmentation uses SAM IoU, detection uses matched IoU, and depth uses AbsRel. For Cosmos CV4 we report the nearest bitrate operating point.}
\label{tab:isorate_main}
\begin{tabular}{lcccc}
\toprule
Task & System & bpp & Metric & Value \\
\midrule
\multirow{4}{*}{Segmentation}
& Cosmos DV4 & 0.0833 & IoU $\uparrow$ & 0.677 \\
& PAT-VCM Det-Aux+Seg-Aux & 0.1242 & IoU $\uparrow$ & 0.687\\
& PAT-VCM Det-Aux+Seg-Aux+FG+BG & 0.1242 &  IoU $\uparrow$ & \textbf{0.764} \\
& Cosmos CV4 (nearest higher-rate) & 0.1671 & IoU $\uparrow$ & 0.509 \\
& MPA q=1 & 0.0908 & IoU $\uparrow$ & 0.739 \\
\midrule
\multirow{3}{*}{Detection}
& Cosmos DV4 & 0.0833 & matched IoU $\uparrow$ & 0.475 \\
& PAT-VCM Det-Aux & 0.1007 & matched IoU $\uparrow$ & \textbf{0.512} \\
& Cosmos CV4 (nearest higher-rate) & 0.167 & matched IoU $\uparrow$ & 0.031 \\
\midrule
\multirow{3}{*}{Depth}
& Cosmos DV4  & 0.0833 & AbsRel $\downarrow$ & 2.296 \\
& PAT-VCM Det-Aux + Depth-Aux & 0.1007 & AbsRel $\downarrow$ & \textbf{2.412} \\
& Cosmos CV4 (nearest higher-rate) & 0.1671 & AbsRel $\downarrow$ & 19.86 \\
\bottomrule
\end{tabular}\vspace{-.5em}
\end{table*}

\subsection{Inference Cost and Bitrate}

Inference time was measured on DAVIS-2017 val at \(512\times512\) resolution using a single NVIDIA L40S GPU (48 GB), with batch size 1.  Cosmos baseline runs at \(101.6\) ms per 9-frame clip (\(29.7\) ms encode, \(71.9\) ms decode). The full encoder-side and decoder-size PAT-VCM pipelines with Det-Aux + Seg-Aux take \(640.0\) ms and \(692.5\) ms per clip, respectively. This corresponds to about \(14.1\) fps for the encoder and \(13.0\) fps for the decoder. With adaptive text tokens, the expected runtime increases only modestly, to about \(704\) ms per clip on the encoder side and \(733\) ms per clip on the decoder side.  On the rate side, the baseline Cosmos stream operates at \(0.0833\) bpp (\(196{,}608\) bits per 9-frame clip). Det-Aux raises this to \(0.1007\) bpp, and the combined Det-Aux + Seg-Aux raises this to \(0.1242\) bpp. Prompt and text-based token streams add negligible bitrate by comparison. Full latency and bitrate details are given in the appendix.~\ref{ap:bitratecost}.

\subsection{Limitations}
The current PAT-VCM design is most effective when the downstream task can benefit from local ROI refinement or compact task-level semantic outputs. Tasks whose predictions depend more strongly on spatial derivatives or cross-boundary consistency, such as surface normals under the current ROI residual formulation, show much smaller gains. Extending the framework with smoother block-based refinement or low-resolution full-frame residuals is a natural direction for dense geometric tasks. Also, later task-specific branches depend on the upstream localization stage. If an object is completely missed by the shared detection-oriented branch, subsequent ROI refinement and prompt tokens cannot recover it. Extending the framework with fallback mechanisms such as global proposal streams or low-rate full-frame cues is an important direction for future work.

In addition, although PAT-VCM avoids retraining a separate end-to-end codec for every downstream model, task-specific auxiliary branches still need task-specific supervision. Extending the framework toward more unified auxiliary learning across tasks is an important direction for future work. 

\section{Conclusion}

We presented PAT-VCM, a plug-and-play auxiliary-token framework for video coding for machines. PAT-VCM keeps a shared baseline compressed stream and augments it with lightweight task-aware auxiliary tokens, rather than retraining a separate full codec for each downstream task and model. This reduces coupling to any single end task and makes the framework easier to extend across tasks and model updates.

Our experiments show that this design is effective across different downstream tasks. A shared detection-oriented auxiliary branch provides a reusable first refinement. On top of it, task-specific visual branches improve segmentation and depth estimation, while prompt/control tokens provide further segmentation gains at negligible bitrate. Overall, the results support a simple conclusion: a shared compressed representation, combined with lightweight auxiliary tokens, provides a practical and flexible alternative to tightly task-specific VCM design.

\bibliographystyle{splncs04}
\bibliography{ref}

\clearpage
\appendix

\section{Additional Detection Analysis}\label{ap:detection}

\subsection{Cross-Dataset Detection Recall}

Table~\ref{tab:det_cross_dataset} reports detection recall for the shared detection-oriented branch across four datasets. Det-Aux is trained on YouTube-VIS 2021 \cite{cheng2021m2fvis} and evaluated zero-shot on DAVIS, MOSE, and VIPSeg. The largest absolute gain appears on DAVIS, where the compression gap is also the largest.

\begin{table*}[htb]
\centering
\caption{Cross-dataset detection recall for the shared detection-oriented branch. Higher is better.}
\label{tab:det_cross_dataset}
\begin{tabular}{lccccc}
\toprule
Dataset & No. GT & Cosmos & + Det-Aux & Orig & \% gap closed \\
\midrule
DAVIS-2017 val & 59  & 0.390 & 0.475 & 0.610 & 39\% \\
MOSE val       & 386 & 0.233 & 0.249 & 0.267 & 47\% \\
VIPSeg val     & 528 & 0.127 & 0.133 & 0.161 & 18\% \\
\bottomrule
\end{tabular}
\end{table*}

\subsection{Matched-Protocol DAVIS Detection}

For direct comparability with the rate-matched evaluation in the main paper, we also report a matched-protocol DAVIS re-evaluation based on Deformable DETR predictions on the uncompressed source as pseudo-ground truth. Table~\ref{tab:det_davis_matched} shows that Det-Aux improves both matched detection IoU and Recall@0.5.

\begin{table}[htb]
\centering
\caption{Matched-protocol detection results on DAVIS. Higher is better.}
\label{tab:det_davis_matched}
\begin{tabular}{lcc}
\toprule
Method & Matched IoU & Recall@0.5 \\
\midrule
Cosmos & 0.475 & 0.516 \\
+ Det-Aux & 0.512 & 0.549 \\
\bottomrule
\end{tabular}
\end{table}

\paragraph{Discussion.}
The cross-dataset recall table and the matched-protocol DAVIS table use related but different evaluation protocols, so the gains should not be compared numerically across tables. The first measures dataset-level recall against the uncompressed source as oracle, while the second measures matched detection quality under the same protocol used in the rate-matched analysis. Together, they show that Det-Aux consistently improves detection and serves as a useful first refinement for later task-specific branches.

\section{Additional Segmentation Analysis}\label{ap:segmentation}

\subsection{Adaptive Text Tokens for Hard-Case}\label{ap:seg_texttoken}

The goal of this experiment is not to replace the main visual and prompt-based segmentation pipeline, but to test whether semantic side information can help in cases where the compressed visual signal is already too degraded for reliable mask prediction.

\paragraph{Setup.}
We evaluate on MOSE val (\(311\) videos, \(377\) ROI instances). The baseline is Cosmos+Seg-Aux. At the encoder, where the uncompressed source frame is available, we generate a short caption for each ROI crop using LLaVA-NeXT-Mistral-7B \cite{liu2023improved}, and encode the caption into a text token stream (about \(152\) bits per ROI). We then run SAM in multimask mode on the compressed frame and use CLIP ViT-B/32 \cite{radford2021clip} to score the candidate masks against the transmitted caption, selecting the mask whose masked crop is most compatible with the text. To understand where text is useful, we partition ROIs by the baseline Cosmos+Seg-Aux IoU into three bins: hard (\(<0.30\)), medium (\([0.30,0.75)\)), and easy (\(\geq 0.75\)).

\paragraph{Hard-case gains.}
Table~\ref{tab:text_token_seg} shows that text tokens are highly effective on hard cases. On the hard bin, the IoU improves from \(0.098\) to \(0.246\), a gain of \(+0.149\). However, applying text tokens uniformly to all ROIs slightly hurts the medium and easy bins, so the overall IoU decreases from \(0.682\) to \(0.652\). This indicates that text is most useful when the compressed visual evidence is already poor, but is less helpful once the visual segmentation is already reliable.

\begin{table}[htb]
\centering
\caption{Text-token segmentation on MOSE val. Metric is mean IoU.}
\label{tab:text_token_seg}
\begin{tabular}{lcccc}
\toprule
Bin & No. instances & Cosmos+SAM & + Text token & \(\Delta\) \\
\midrule
Hard (\(<0.30\)) & 26  & 0.098 & 0.246 & +0.149 \\
Medium (\([0.30,0.75)\)) & 160 & 0.592 & 0.586 & -0.006 \\
Easy (\(\geq 0.75\)) & 191 & 0.853 & 0.836 & -0.017 \\
\midrule
Overall & 377 & 0.682 & 0.652 & -0.030 \\
\bottomrule
\end{tabular}
\end{table}

\begin{figure}[t]
\centering
\includegraphics[width=\textwidth]{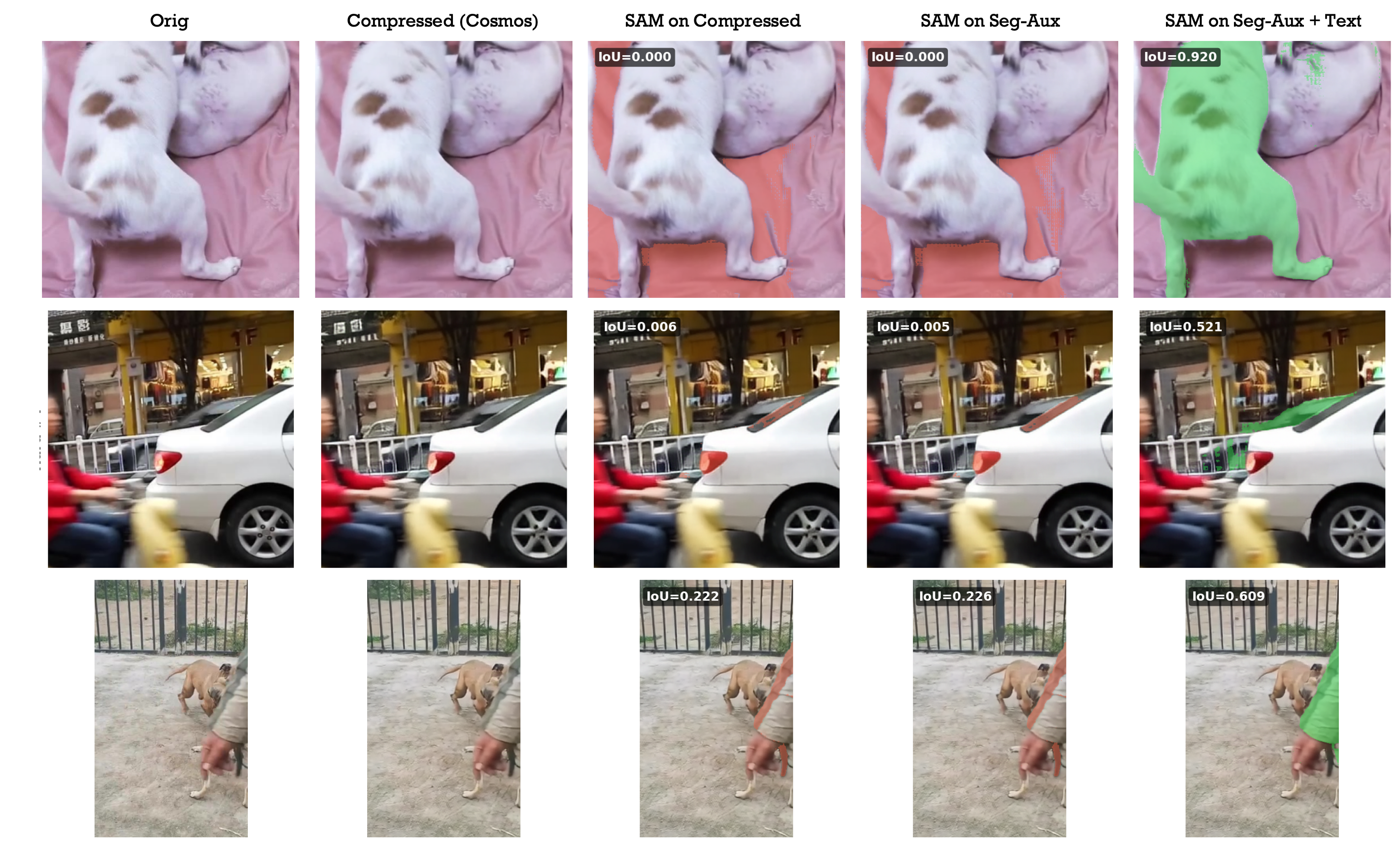}\vspace{-.5em}
\caption{Qualitative hard-case segmentation examples with text-based semantic tokens. From left to right: original frame, compressed reconstruction, segmentation on the compressed reconstruction, and result after adding text tokens. In these cases, the compressed visual signal is too weak for reliable mask selection, while the encoder-side text token provides semantic guidance that helps recover the correct object mask.}
\label{fig:seg_text_qual}\vspace{-1em}
\end{figure}

Figure~\ref{fig:seg_text_qual} shows representative hard cases for text-based semantic tokens. In these examples, the compressed reconstruction does not preserve enough local detail for reliable segmentation, and the baseline SAM prediction tends to select the wrong mask or miss part of the object. Adding the text token helps disambiguate the object identity and improves mask selection. This is consistent with the quantitative result that text tokens are most useful on hard cases, while their benefit is limited on medium and easy cases.

\paragraph{Adaptive transmission.}
Because text helps only a small subset of difficult ROIs, an adaptive policy is preferable to uniform transmission. We therefore consider a simple policy that transmits text only for hard cases. Table~\ref{tab:text_token_policy} shows that this adaptive policy preserves the hard-bin gain while avoiding the regression on easier cases. It improves the overall IoU from \(0.682\) to \(0.692\), while requiring only \(10.4\) bits per ROI on average, compared with \(154\) bits per ROI for uniform text transmission. The corresponding bitrate increase over the Cosmos baseline is negligible.

\begin{table}[htb]
\centering
\caption{Uniform versus adaptive text-token policies on MOSE val.}
\label{tab:text_token_policy}
\begin{tabular}{lccccc}
\toprule
Policy & Avg bits/ROI & Avg bits/clip & \(\Delta\) bpp & Hard-bin IoU & Overall IoU \\
\midrule
No text & 0   & 0    & 0         & 0.098 & 0.682 \\
Uniform text & 154 & 231  & 0.000088 & 0.246 & 0.652 \\
Adaptive text & 10.4 & 15.6 & 0.000006 & 0.246 & 0.692 \\
\bottomrule
\end{tabular}
\end{table}

\paragraph{Discussion.}
This experiment supports the broader view of PAT-VCM that different auxiliary token types are useful in different scenarios. Visual residual tokens improve the general image evidence, prompt tokens improve decoder interaction, and text tokens can recover performance in hard cases where local pixel recovery is no longer sufficient. The result also suggests that the encoder should adaptively choose which type of auxiliary information to transmit for each ROI, rather than applying a single token type uniformly.

\subsection{Token-Type Ablation}\label{ap:seg_tokenablation}

To isolate the role of different auxiliary token types, we compare visual-only refinement, prompt-only refinement, and their combination on DAVIS. Table~\ref{tab:seg_token_ablation_app} shows that prompt-only refinement on top of Det-Aux already gives a much larger gain than Seg-Aux alone, while the combination performs best. This supports the view that visual refinement and prompt refinement address different bottlenecks and are complementary rather than redundant.

\begin{table}[h]
\centering
\caption{Token-type ablation for segmentation on DAVIS. Metric is mean IoU.}
\label{tab:seg_token_ablation_app}
\begin{tabular}{lcc}
\toprule
Variant & Mean IoU & Gain over Cosmos \\
\midrule
Cosmos & 0.677 & +0.000 \\
+ Det-Aux & 0.687 & +0.010 \\
+ Det-Aux + 1-pt & 0.740 & +0.063 \\
+ Det-Aux + Seg-Aux & 0.710 & +0.033 \\
+ Det-Aux + Seg-Aux + 1-pt & 0.751 & +0.074 \\
Orig & 0.774 & +0.097 \\
\bottomrule
\end{tabular}
\end{table}

\subsection{Difficulty-Bin Analysis}\label{ap:seg_difficultbin}

We further partition DAVIS instances by Cosmos baseline IoU quality: easy  \(\geq 0.75\), medium \(0.3\) -- \(0.75\), hard \(< 0.30\). Table~\ref{tab:seg_difficulty_app} shows that prompt refinement is especially effective on medium and hard cases, while Seg-Aux contributes more moderately. On easy cases, the full system remains stable and still improves average quality.

\begin{table*}[h]
\centering
\caption{Segmentation performance by difficulty bin on DAVIS. Bin are defined by Cosmos IoU.}
\label{tab:seg_difficulty_app}
\small
\begin{tabular}{lccccc}
\toprule
Difficulty & No. instance & Cosmos & + Det-Aux & + Det-Aux + Seg-Aux & + Det-Aux + Seg-Aux + 1-pt \\
\midrule
easy & 23 & 0.8742 & 0.8780 & 0.8715 & 0.9010 \\
medium & 29 & 0.6281 & 0.6443 & 0.6823 & 0.7236 \\
hard & 6 & 0.1661 & 0.1551 & 0.2208 & 0.3382 \\
\bottomrule
\end{tabular}
\end{table*}

\subsection{ROI-Size Sensitivity}\label{ap:seg_roisensitivity}

Table~\ref{tab:seg_roi_size_app} shows segmentation performance as a function of ROI size on DAVIS. Prompt refinement is most effective on medium and large ROIs, while gains on small ROIs are more modest.

\begin{table}[h]
\centering
\caption{Segmentation ROI-size sensitivity on DAVIS. Metric is mean IoU.}
\label{tab:seg_roi_size_app}
\small
\begin{tabular}{lccccc}
\toprule
Size bin & No. instance & Cosmos & + Det-Aux & + Det-Aux + Seg-Aux & + Det-Aux + Seg-Aux + 1-pt \\
\midrule
Small  & 20 & 0.6010 & 0.5953 & 0.6131 & 0.6278 \\
Medium & 32 & 0.7263 & 0.7366 & 0.7611 & 0.8209 \\
Large  & 6  & 0.6762 & 0.7222 & 0.7562 & 0.8188 \\
\bottomrule
\end{tabular}
\end{table}

\subsection{No-Harm Retention on Easy Cases}\label{ap:seg_noharm}

We also check whether the auxiliary branches damage already easy cases. On DAVIS easy cases with Cosmos IoU \(\geq 0.75\) (\(n=23\)), Seg-Aux harms \(6/23\) cases by more than \(0.01\) IoU, while Seg-Aux plus prompt harms only \(1/23\). The average IoU change on easy cases is still positive at \(+0.0267\).

\section{Additional Depth Analysis} \label{ap:depthestimation}

\subsection{Difficulty-bin analysis}\label{ap:depth_difficultbin}
As with segmentation, the depth gains are concentrated on difficult cases. Table~\ref{tab:depth_difficulty} partitions DAVIS instances by Cosmos baseline ROI AbsRel: easy \(<0.05\), medium \(0.05\) -- \(0.20\), hard \(\geq 0.20\). The auxiliary streams have little effect on already easy cases, but Depth-Aux provides large gains on hard examples.

\begin{table}[h]
\centering
\caption{Depth performance by difficulty bin on DAVIS. AbsRel is lower-is-better.}
\label{tab:depth_difficulty}
\begin{tabular}{lcccc}
\toprule
Difficulty & No. instance & Cosmos & + Det-Aux & + Det-Aux + Depth-Aux \\
\midrule
easy & 4 & 0.0341 & 0.0332 & 0.0394 \\
medium & 31 & 0.1243 & 0.1270 & 0.1320 \\
hard  & 23 & 4.2539 & 3.5733 & 2.2300 \\
\bottomrule
\end{tabular}
\end{table}

\subsection{Depth Baseline and Shared Det-Aux Characterization}\label{ap:depth_baselineshared}
Table~\ref{tab:depth_baseline} characterizes depth estimation from the compressed baseline and after the shared Det-Aux refinement, before introducing Depth-Aux. The main pattern is that DAVIS is much more affected by compression than VIPSeg. On DAVIS, the error is especially high inside ROI regions, where Cosmos gives an AbsRel of 3.764 compared with 2.576 on the full frame. The shared Det-Aux branch reduces this error on DAVIS, improving ROI AbsRel from 3.764 to 3.057 and full-frame AbsRel from 2.576 to 2.151, but a substantial gap remains. In contrast, VIPSeg is already much more stable under compression, and Det-Aux changes the depth metrics only slightly. These results show that the depth gap is concentrated mainly on DAVIS, especially in ROI regions, which motivates a task-specific depth refinement branch.

\begin{table*}[h]
\centering
\caption{Baseline depth characterization. Metrics are reported after per-frame scale--shift alignment. Lower AbsRel and RMSE are better; higher \(\delta<1.25\) is better.}
\label{tab:depth_baseline}
\begin{tabular}{llccc}
\toprule
Dataset & Variant & AbsRel & RMSE & \(\delta<1.25\) \\
\midrule
DAVIS full-frame & Cosmos & 2.576 & 0.315 & 0.810 \\
DAVIS full-frame & + Det-Aux & 2.151 & 0.310 & 0.807 \\
DAVIS ROI-only   & Cosmos & 3.764 & 0.364 & 0.828 \\
DAVIS ROI-only   & + Det-Aux & 3.057 & 0.357 & 0.834 \\
\midrule
VIPSeg full-frame & Cosmos & 0.484 & 0.187 & 0.893 \\
VIPSeg full-frame & + Det-Aux & 0.476 & 0.187 & 0.895 \\
VIPSeg ROI-only   & Cosmos & 0.485 & 0.188 & 0.892 \\
VIPSeg ROI-only   & + Det-Aux & 0.477 & 0.187 & 0.894 \\
\bottomrule
\end{tabular}
\end{table*}

\section{Inference Cost and Bitrate}\label{ap:bitratecost}

\subsection{End-to-End Runtime by Task}

Table~\ref{tab:e2e_runtime_task} reports overall runtime for the main pipelines on a single NVIDIA L40S GPU. We report encoder-side and decoder-side runtime separately, since PAT-VCM uses asymmetric processing at the two ends. 

\begin{table*}[htb]
\centering
\caption{Overall runtime by task and pipeline on a single NVIDIA L40S GPU.}
\label{tab:e2e_runtime_task}
\small
\begin{tabular}{lcc}
\toprule
 Pipeline & Latency (ms / clip) & Throughput \\
\midrule
 Cosmos & 101.6 & 88.6 fps \\
\midrule
 Encoder: Cosmos + DETR + Det-Aux & 488.3 & 18.4 fps \\
 Decoder: Cosmos + Det-Aux + DETR & 530.5 & 17.0 fps \\
\midrule
 Encoder: Cosmos + DETR + Det-Aux + Seg-Aux & 640.0 & 14.1 fps \\
 Decoder: Cosmos + Det-Aux + Seg-Aux + SAM & 692.5 & 13.0 fps \\
 Encoder: above + adaptive LLaVA text & 704.0 & 12.8 fps \\
 Decoder: above + adaptive multimask SAM + CLIP reranking & 733.0 & 12.3 fps \\
\midrule
 Encoder: Cosmos encode + DETR + Det-Aux + Depth-Aux & 665.4 & 13.5 fps \\
 Decoder: Cosmos decode + Det-Aux + Depth-Aux + Depth Anything & 1242.2 & 7.2 fps \\
\bottomrule
\end{tabular}
\end{table*}



\subsection{Bitrate Accounting}

We report bitrate using the standard convention
\[
\mathrm{bpp}=\frac{\text{total bits per clip}}{T\times H\times W},
\]
with \(T=9\) and \(H=W=512\). Table~\ref{tab:bitrate_cost} gives the detailed bitrate information. The main observation is that the visual auxiliary streams account for the meaningful bitrate increase, while prompt and text-based side information are negligible by comparison. Det-Aux adds about \(0.017\) bpp over the baseline, and the combined Det-Aux + Seg-Aux increases the operating point to \(0.1242\) bpp. By contrast, text and codebook tokens does not change the bitrate in a meaningful way.

\begin{table*}[htb]
\centering
\caption{Bitrate counts for a 9-frame \(512\times512\) clip.}
\label{tab:bitrate_cost}
\begin{tabular}{lcccc}
\toprule
Variant  & Extra & Total & Bits / frame & bpp \\
\midrule
Cosmos DV baseline & 0 & 196,608 & 21,845 & 0.08333 \\
+ Det-Aux (measured ROI-masked)  & 40,960 & 237,568 & 26,396 & 0.10069 \\
+ Det-Aux (full-frame upper bound)  & 147,456 & 344,064 & 38,229 & 0.14583 \\
+ Det-Aux + Seg-Aux & 96,460 & 293,068 & 32,563 & 0.12422 \\
+ Text + box  & 228 & 196,836 & 21,871 & 0.08343 \\
+ 1-pt  & 5 & 196,615 & 21,846 & 0.08334 \\
+ FG + BG  & 10 & 196,615 & 21,846 & 0.08334 \\
\bottomrule
\end{tabular}
\end{table*}

\section{Rate-Matched Evaluation}\label{ap:isorate}

This appendix reports additional rate-matched results for PAT-VCM. The goal is to separate the effect of the auxiliary architecture from the effect of additional bits alone. We compare PAT-VCM against a frozen Cosmos CV4 quantization sweep within the same model family, and against MPA~\cite{liu2024mpa} for segmentation on DAVIS. The CV4 baseline uses the continuous-latent variant of Cosmos, whose latent is scalar-quantized at different bit depths and decoded with the same frozen model. 

\subsection{Comparison with Rate-Matched Cosmos CV4}

Table~\ref{tab:isorate_davis_main} summarizes the main DAVIS comparisons against the nearest higher-rate Cosmos CV4 operating point from the quantization sweep. For segmentation, the full PAT-VCM pipeline (Det-Aux + Seg-Aux + FG+BG) substantially outperforms the nearest higher-rate Cosmos CV4 baseline. For detection, Det-Aux also clearly outperforms Cosmos CV4 at comparable rate. For depth, the Cosmos CV4 sweep remains much worse than the DAVIS Det-Aux operating point under the same evaluation protocol. These results indicate that the gains are not explained by added bits alone, but by how the auxiliary information is allocated and used. 

\begin{table*}[t]
\centering
\caption{Rate-matched comparison on DAVIS against the Cosmos CV4 quantization sweep. For CV4, we report the nearest higher-rate operating point from the sweep.}
\label{tab:isorate_davis_main}
\begin{tabular}{lcccc}
\toprule
Task & System & bpp & Metric & Value \\
\midrule
\multirow{3}{*}{Segmentation}
& Cosmos DV4 baseline & 0.0833 & IoU \(\uparrow\) & 0.6761 \\
& PAT-VCM (Det-Aux + Seg-Aux + FG+BG) & 0.1242 & IoU \(\uparrow\) & \textbf{0.7640} \\
& Cosmos CV4 (nearest higher-rate) & 0.1671 & IoU \(\uparrow\) & 0.5087 \\
\midrule
\multirow{3}{*}{Detection}
& Cosmos DV4 baseline & 0.0833 & matched IoU \(\uparrow\) & 0.4751 \\
& PAT-VCM (Det-Aux) & 0.1007 & matched IoU \(\uparrow\) & \textbf{0.5116} \\
& Cosmos CV4 (nearest higher-rate) & 0.1671 & matched IoU \(\uparrow\) & 0.0309 \\
\midrule
\multirow{3}{*}{Depth}
& Cosmos DV4 baseline & 0.0833 & AbsRel \(\downarrow\) & 2.296 \\
& PAT-VCM (Det-Aux) & 0.1007 & AbsRel \(\downarrow\) & 2.412 \\
& Cosmos CV4 (nearest higher-rate) & 0.1671 & AbsRel \(\downarrow\) & 19.86 \\
\bottomrule
\end{tabular}
\end{table*}

For completeness, Tables~\ref{tab:isorate_seg_sweep}, \ref{tab:isorate_det_sweep}, and \ref{tab:isorate_depth_sweep} report the full Cosmos CV4 sweep on DAVIS. The segmentation results show that Cosmos CV4 needs much higher bitrate to approach the quality of the PAT-VCM segmentation pipeline. Detection is particularly sensitive to Cosmos CV4 quantization, with severe degradation at low rates. Depth also degrades sharply under Cosmos CV4 quantization.

\begin{table}[htb]
\centering
\caption{Full DAVIS segmentation sweep for Cosmos CV4.}
\label{tab:isorate_seg_sweep}
\begin{tabular}{lcc}
\toprule
System & bpp & IoU \(\uparrow\) \\
\midrule
Cosmos DV4 & 0.0833 & 0.6761 \\
Cosmos DV4 + Det-Aux + Seg-Aux & 0.1242 & 0.6874 \\
Cosmos CV4 N=1 & 0.0838 & 0.5460 \\
Cosmos CV4 N=2 & 0.1671 & 0.5087 \\
Cosmos CV4 N=3 & 0.2504 & 0.6328 \\
Cosmos CV4 N=10 & 0.8338 & 0.6764 \\
\bottomrule
\end{tabular}
\end{table}

\begin{table}[htb]
\centering
\caption{Full DAVIS detection sweep for Cosmos CV4.}
\label{tab:isorate_det_sweep}
\begin{tabular}{lcc}
\toprule
System & bpp & matched IoU \(\uparrow\) \\
\midrule
Cosmos DV4 & 0.0833 & 0.4751 \\
Cosmos DV4 + Det-Aux & 0.1007 & 0.5116 \\
Cosmos CV4 N=1 & 0.0838 & 0.0093 \\
Cosmos CV4 N=2 & 0.1671 & 0.0309 \\
Cosmos CV4 N=3 & 0.2504 & 0.2717 \\
Cosmos CV4 N=10 & 0.8338 & 0.5229 \\
\bottomrule
\end{tabular}
\end{table}

\begin{table}[htb]
\centering
\caption{Full DAVIS depth sweep for Cosmos CV4.}
\label{tab:isorate_depth_sweep}
\begin{tabular}{lcc}
\toprule
System & bpp & AbsRel \(\downarrow\) \\
\midrule
Cosmos DV4 & 0.0833 & 2.296 \\
Cosmos DV4 + Det-Aux & 0.1007 & 2.412 \\
Cosmos CV4 N=1 & 0.0838 & 23.55 \\
Cosmos CV4 N=2 & 0.1671 & 19.86 \\
Cosmos CV4 N=3 & 0.2504 & 2.361 \\
Cosmos CV4 N=10 & 0.8338 & 1.448 \\
\bottomrule
\end{tabular}
\end{table}




\subsection{Additional Comparison with MPA}

We also compare against MPA~\cite{liu2024mpa}, a recent joint human--machine image codec. Since MPA is an image codec, we apply it per-frame to the DAVIS middle frame and evaluate it under the same SAM-based segmentation protocol used elsewhere in the paper. Table~\ref{tab:mpa_davis_app} reports the results. Under this matched protocol, the full PAT-VCM segmentation pipeline (Det-Aux + Seg-Aux + FG+BG) exceeds MPA on DAVIS segmentation while operating at a comparable rate. The visual-only PAT-VCM stack is weaker, which highlights the importance of prompt/control tokens in the full framework.

\begin{table}[htb]
\centering
\caption{Comparison with MPA on DAVIS under matched per-frame SAM evaluation.}
\label{tab:mpa_davis_app}
\begin{tabular}{lcc}
\toprule
System & bpp & SAM IoU \(\uparrow\) \\
\midrule
MPA q=1 & 0.0908 & 0.7387 \\
MPA q=2 & 0.1320 & 0.7315 \\
MPA q=3 & 0.1914 & 0.7454 \\
MPA q=8 & 1.0189 & 0.7656 \\
Cosmos baseline & 0.0833 & 0.6761 \\
Det-Aux + Seg-Aux & 0.1242 & 0.6874 \\
Det-Aux + Seg-Aux + FG+BG & 0.1242 & \textbf{0.7640} \\
Uncompressed & -- & 0.7416 \\
\bottomrule
\end{tabular}
\end{table}

\paragraph{Discussion.}
The rate-matched comparison with Cosmos CV4 shows that the auxiliary-token gains are not explained by added bitrate alone. The comparison with MPA highlights a different point: PAT-VCM is not merely a stronger image codec, but a framework that combines a frozen video codec with lightweight task-aware auxiliary streams. The full segmentation pipeline benefits substantially from the combination of visual refinement and prompt/control tokens, which is not captured by the visual-only stack. 

\section{Additional Task Study: Semantic Recognition}\label{ap:recognitiontask}

\paragraph{Visual auxiliary branch.}
A recognition-oriented visual branch can be trained to preserve CLIP features on object ROIs. In this case, the auxiliary branch refines the crop so that the CLIP embedding of the refined crop remains close to that of the original crop. This gives a modest but consistent gain over the baseline.

\paragraph{Semantic tokens.}
Recognition also uses a more compact semantic-token form. This should be viewed as an upper-end case of the framework: given an object ROI from the upstream localization stage, the encoder transmits a lightweight task-level semantic result directly, rather than further refining pixels. In the current implementation, the encoder applies the frozen CLIP vision encoder to the ROI crop from the original frame and compares the resulting feature with a fixed set of text embeddings for the candidate object classes. The predicted class is then encoded as a discrete class-label token and transmitted. Since the current setup uses \(80\) candidate classes, the label can be represented with \(7\) bits per ROI. At the decoder, this token is interpreted directly as the recognized class for that ROI, without requiring CLIP inference on the compressed crop. The result therefore measures semantic preservation conditioned on localization, not joint detection and recognition. Table~\ref{tab:semantic_token} shows that this semantic token substantially outperforms visual refinement while requiring only \(7\) bits per ROI. This suggests that when the downstream target is already semantic, compact task-level tokens can preserve task performance much more efficiently than additional pixel refinement.

\begin{table}[t]
\centering
\caption{Semantic-token results for CLIP recognition on DAVIS.}\vspace{-.3em}
\label{tab:semantic_token}
\begin{tabular}{lccc}
\toprule
Method & Class agreement (\%) $\uparrow$ & Bits / ROI & No. instances \\
\midrule
Cosmos (no token) & 64.9 & 0 & 57 \\
+ Det-Aux (no token) & 63.2 & 0 & 57 \\
+ Det-Aux + Class-label token & 100.0 & 7 & 57 \\
\bottomrule
\end{tabular}
\end{table}

\section{Additional Task Study: Surface Normals}\label{ap:surfacenormal}

This appendix reports an additional task study that helps clarify the current scope of PAT-VCM under the ROI-based residual design. Unlike segmentation and depth estimation, surface normal estimation shows only marginal gains.

We evaluate surface normals as a secondary geometric task. The normal-specific auxiliary branch uses the same auxiliary architecture as the main task-specific branches,
\[
\texttt{AuxEncoder-256 + FSQ[8,8,8,8] + ResidualDecoder-128},
\]
and is trained on ROI regions with a cosine similarity loss on surface normals. In our setup, normals are derived from depth predictions through spatial gradients, so the task depends on local surface orientation rather than absolute depth values. 

\paragraph{Results.}
Table~\ref{tab:appendix_normal} summarizes the ROI normal results on DAVIS. The normal-specific branch gives only a marginal improvement over the shared Det-Aux baseline. In addition, the depth-specific branch slightly degrades normal quality, even though it improves absolute depth estimation.

\begin{table}[h]
\centering
\caption{Surface normal estimation on DAVIS ROI regions. MAE is mean angular error in degrees; lower is better.}
\label{tab:appendix_normal}
\begin{tabular}{lc}
\toprule
Model & ROI MAE\(^\circ\) \\
\midrule
Cosmos & 5.13 \\
+ Det-Aux & 5.07 \\
+ Normal-Aux & \textbf{5.06} \\
+ Depth-Aux & 5.17 \\
\bottomrule
\end{tabular}
\end{table}

\paragraph{Discussion.}
These results suggest that the current ROI residual design is less effective for gradient-derived targets than for absolute per-pixel predictions such as depth. A likely reason is that surface normals depend on spatial derivatives, so small discontinuities at ROI boundaries may have limited effect on depth values but noticeably affect local orientation. 

\section{Additional Task Study: Pose Estimation}\label{ap:poseestimation}

We also evaluate human pose estimation as an additional structured prediction task. The downstream model is a frozen Keypoint R-CNN with a ResNet50-FPN backbone. In the visual-auxiliary setting, the pose branch uses the same auxiliary architecture as the segmentation and depth branches and is trained by feature distillation on the frozen Keypoint R-CNN backbone within person ROIs. In addition to this visual branch, pose also admits a semantic-token form in which compact keypoint outputs are transmitted directly. The detailed architecture and training setup follow the same shared auxiliary design used in the main tasks.

\paragraph{Baseline compression gap.}
Pose estimation has a clear compression gap. The shared Det-Aux branch does not consistently improve this task. On DAVIS, the baseline error corresponds to about \(5.5\%\) of the frame width, indicating that compression noticeably affects fine-grained joint localization.

\paragraph{Semantic skeleton tokens.}
For pose, we further evaluate a semantic-token form in which quantized keypoints are transmitted directly. In the current implementation, the skeleton token uses an \(8\times 8\) quantization grid and requires \(102\) bits per person. Table~\ref{tab:appendix_pose} summarizes the DAVIS results. The skeleton token reduces MKE from \(33.62\) pixels for the Cosmos baseline to \(7.33\) pixels, corresponding to a \(78\%\) reduction. Although this token is larger than the class-label token used for recognition, it is still negligible compared with the ROI pixel auxiliary stream.

\begin{table}[h]
\centering
\caption{Pose estimation on DAVIS. MKE is mean keypoint error in pixels; lower is better.}
\label{tab:appendix_pose}
\begin{tabular}{lcc}
\toprule
Method & MKE (px)\(\downarrow\) & Bits / person \\
\midrule
Cosmos (no token) & 33.62 & 0 \\
+ Det-Aux (no token) & 31.99 & 0 \\
+ Skeleton token & \textbf{7.33} & 102 \\
\bottomrule
\end{tabular}
\end{table}

\paragraph{Discussion.}
The pose results are consistent with the broader PAT-VCM framework. When the downstream target itself is structured and compact, transmitting a semantic token can be far more effective than relying on additional pixel refinement alone. At the same time, the larger token size relative to recognition reflects the fact that pose must encode a richer structured output than a single class label. Even so, the bitrate remains very small compared with the pixel auxiliary stream. 

\begin{table*}[t]
\centering
\caption{Training datasets for auxiliary branches.}
\label{tab:train_eval_matrix}
\begin{tabular}{llll}
\toprule
Branch & Training dataset & In-domain held-out \\
\midrule
Det-Aux & YTVIS 2021 splits 0--6 & YTVIS split\_7  \\
Seg-Aux feat-distill & YTVIS 2021 splits 0--6 & YTVIS split\_7  \\
Seg-Aux SAM-aware & MOSE train & MOSE val  \\
Depth-Aux & VIPSeg train & VIPSeg val  \\
Normal-Aux & VIPSeg train & --  \\
Pose-Aux (primary) & VIPSeg train & -- \\
CLIP-Aux & MOSE train & --  \\
\bottomrule
\end{tabular}
\end{table*}

\section{Auxiliary Training Details}\label{ap:trainingobjectives}

All task models and the baseline Cosmos codec remain frozen during auxiliary training. Only the auxiliary encoder, FSQ module, and residual decoder of each branch are updated.

\subsection{Det-Aux Loss}

Det-Aux is trained with a detection-oriented distillation objective
\[
\mathcal{L}_{\mathrm{det}}
=
\mathcal{L}_{\mathrm{feat}}
+ 0.03\,\mathcal{L}_{\mathrm{enc}}
+ \mathcal{L}_{\mathrm{L1}},
\]
where \(\mathcal{L}_{\mathrm{feat}}\) is the weighted MSE between frozen Deformable DETR backbone features on the refined and original frames over levels \(\{0,1,2\}\) with weights \(\{0.25,1.0,3.0\}\), \(\mathcal{L}_{\mathrm{enc}}\) is the analogous MSE on the DETR transformer-encoder fused feature map, and \(\mathcal{L}_{\mathrm{L1}}\) is the mean absolute pixel error inside the Stage-1 ROI regions. We do not use DETR detection-head losses such as box regression or classification losses.

\subsection{Seg-Aux Loss}

Seg-Aux is trained with a segmentation-oriented feature-distillation objective
\[
\mathcal{L}_{\mathrm{seg}}
=
\mathcal{L}_{\mathrm{encfeat}}
+ 0.03\,\mathcal{L}_{\mathrm{pixeldec}}
+ \mathcal{L}_{\mathrm{L1}},
\]
where \(\mathcal{L}_{\mathrm{encfeat}}\) is the ROI-masked weighted MSE between frozen Mask2Former encoder hidden states on the refined and original frames over levels \(\{0,1,2,3\}\) with weights \(\{0.5,0.5,2.0,2.0\}\), \(\mathcal{L}_{\mathrm{pixeldec}}\) is the analogous ROI-masked MSE on the Mask2Former pixel-decoder last hidden state, and \(\mathcal{L}_{\mathrm{L1}}\) is the mean absolute pixel error inside the Stage-2 ROI regions. The ROI mask is downsampled to each feature level so the loss is computed only over object-aligned positions.

\subsection{Depth-Aux and CLIP-Aux Losses}

Depth-Aux is trained with a scale-invariant log-depth loss under frozen Depth Anything V2 predictions together with an ROI-localized L1 regularizer:
\[
\mathcal{L}_{\mathrm{depth}}
=
\mathcal{L}_{\mathrm{sid}}
+ \lambda_{\mathrm{L1}} \mathcal{L}_{\mathrm{L1}}.
\]

CLIP-Aux is trained with a cosine-similarity loss on ROI crops:
\[
\mathcal{L}_{\mathrm{CLIP}}
=
\frac{1}{N}\sum_{i=1}^{N}
\left(1-\cos\!\left(g_{\mathrm{CLIP}}(\hat{x}^{(i)}_{\mathrm{crop}}),\, g_{\mathrm{CLIP}}(x^{(i)}_{\mathrm{crop}})\right)\right)
+ \lambda_{\mathrm{L1}} \mathcal{L}_{\mathrm{L1}},
\]
where \(g_{\mathrm{CLIP}}\) is the frozen CLIP image encoder.

\subsection{Training Datasets}

Table~\ref{tab:train_eval_matrix} summarizes the training datasets for all auxiliary branches. No auxiliary branch is trained on any evaluation video.

\subsection{Common Training Settings}

Unless otherwise noted, auxiliary branches are trained with AdamW, cosine annealing, and mixed-precision training. Detection auxiliary stream uses learning rate \(2\times10^{-4}\). Task-specific auxiliary branches (segmentation, detection, normal, pose) use learning rate \(5\times10^{-5}\). Weight decay is \(10^{-4}\), resolution is \(512\times512\), and each branch has about \(5.5\)M trainable parameters.

\end{document}